\documentclass[runningheads]{llncs}
\usepackage[T1]{fontenc}
\usepackage{graphicx}
\usepackage{hyperref}
\usepackage{amsmath, amssymb}
\usepackage{cite}
\usepackage{soul}
\usepackage{color}
\usepackage{multirow}
\usepackage{subfig}
\usepackage{color}

\usepackage{marvosym} % for email symbol \Letter
\begin{document}
\title{MIFR: A Modality-Invariant and Fair Representation Framework for Skin Disease Classification}
\titlerunning{MIFR: A Modality Invariant Framework for Skin Disease Classification}
	
\author{
    Asonyu Senge Njih\inst{1} \and 
    Yvan Guifo Fodjo\inst{2} \and 
    Vianney Kengne Tchendji\inst{1} \and 
    Jerry Lacmou Zeutouo\inst{3} \and 
    Kerol Djoumessi\inst{4}\textsuperscript{(\Letter)}
    }
	
\authorrunning{Asonyu et al.}
\institute{Department of Mathematics and Computer Science, University of Dschang, Cameroon \\
 %\email{vianney.kengne@univ-dschang.org} 
\and EFREI Research Lab, Université Paris-Panthéon-Assas,  France  \\ %\email{yvan.guifo-fodjo@efrei.fr}
\and MIS, Université de Picardie Jules Verne, France \\
 %\email{jerry.lacmou.zeutouo@u-picardie.fr}
\and Hertie Institute for AI in Brain Health, University of T\"{u}bingen, Germany 
\email{kerol.djoumessi-donteu@uni-tuebingen.de} \\
\url{https://hertie.ai/}} 

\maketitle              % typeset the header of the contribution

\begin{abstract}
    Skin diseases represent a major global public health burden, yet machine learning tools developed to assist in their diagnosis suffer from two critical limitations: reliance on only one modality for diagnosis and systematic performance disparities across skin tones.
    While existing approaches address each challenge separately, this work proposes a modality-invariant framework with fair representation (MIFR) for skin disease classification. 
    The architecture pairs clinical photographs with dermoscopic images using ViT-based encoders, projecting each input into a high-dimensional embedding space via modality-specific projection heads. The resulting model is trained with a five-component multi-objective loss including weighted cross-entropy for classification, confusion and skin-type classification losses for fairness, per-modality supervised contrastive loss for class alignment, and a modality-invariance loss for clinical and dermoscopic modality alignment. Experiments on the HIBA+Derm7pt paired dataset and the external PAD-UFES-20 and ISIC 2019 datasets showed that modality-invariant representation learning %consistently
    provides competitive predictive performance compare to relevant baseline models and competitive fairness on the internal dataset. t-SNE visualizations confirmed that clinical and dermoscopic embeddings of the same disease are geometrically aligned, %and that skin-tone information is effectively suppressed, 
    validating the joint objectives.

\end{abstract}

\section{Introduction}
    Skin diseases affect nearly one-third of the world's population and represent a major global health burden~\cite{karimkhaniGlobalBurdenScabies2017}. Deep learning has brought dermatology to a new level, with CNNs and Vision Transformers matching or even surpassing dermatologists in diagnostic performance~\cite{dosovitskiyImageWorth16x162021, takahashiComparisonVisionTransformers2024}. Yet, two major issues hinder their reliable deployment in real clinical settings. First, skin lesions are captured through heterogeneous imaging modalities, mostly clinical photographs and dermoscopic images, and relying on a single modality can be insufficient for accurate diagnosis~\cite{10.3389/fpubh.2026.1788454, KEMENESS202670}. Second, most public dermatology datasets are dominated by lighter skin tones, causing models to perform significantly worse on darker skin and perpetuating healthcare disparities~\cite{grohEvaluatingDeepNeural2021, daneshjouDisparitiesDermatologyAI2022}. Importantly, most existing fairness‑aware methods are developed for single‑modality clinical images~\cite{duFairDisCoFairerAI2022}, while recent multimodal foundation models do not incorporate fairness constraints~\cite{yanMultimodalVisionFoundation2025}, leaving the intersection of modality invariance and fairness largely unexplored.

    Among fairness‑aware approaches, \textbf{FairDisCo}~\cite{duFairDisCoFairerAI2022} achieves skin‑tone debiasing via disentanglement and contrastive learning, but operates on a single modality. \textbf{FairAdaBN}~\cite{xuFairAdaBNMitigatingUnfairness2023} replaces batch normalization with adaptive batch normalization, but requires sensitive-attribute labels at test time. \textbf{PatchAlign}~\cite{aayushmanFairAccurateSkin2024} aligns image patches with clinical text descriptions to improve fairness, but depends on costly natural‑language annotations and remains single‑modal on images. In contrast, \textbf{PanDerm}~\cite{yanMultimodalVisionFoundation2025} demonstrates the effectiveness of multimodal vision foundation models across clinical, dermoscopic, dermatopathological, and total-body-photography images but it does not address demographic equity. These methods share three limitations: (i) they treat modality invariance and fairness as separate problems; (ii) most, such as FairAdaBN, require sensitive attribute labels at inference; and (iii) cross‑dataset evaluation remains limited.
    
    In this work, we propose a method that overcomes these gaps by: (1) designing a dual‑encoder architecture that jointly processes clinical and dermoscopic images; (2) requiring FST annotations only for an auxiliary training objective, not at inference; (3) introducing a unified multi‑objective loss that simultaneously enforces modality invariance and skin‑tone fairness; and (4) rigorously evaluating cross‑dataset generalization on two external benchmarks, PAD‑UFES‑20 and ISIC 2019, consisting of clinical and dermoscopic images respectively.

\section{Proposed Modality-invariant and Fair Model}
    \subsection{Problem Formulation}
        Let \( \mathcal{D}_{\text{pair}} = \{(\mathbf{X}_i^c, \mathbf{X}_i^d, y_i, s_i)\}_{i=1}^{N} \) denote paired examples for a number of pairs \(N\), where \( \mathbf{X}_i^c \) and \( \mathbf{X}_i^d \) are clinical and dermoscopic images of the same lesion, \( y_i \in \{0,1,..., c\} \) the disease class, and \( s_i \in \{1,\dots,6\} \) the Fitzpatrick Skin Type. Unpaired sets \( \mathcal{D}_c \) and \( \mathcal{D}_d \) are also available. The full training set consists of \( D = \mathcal{D}_{\text{pair}}\cup \mathcal{D}_c\cup \mathcal{D}_d \). We seek encoders \( f_\theta^c \), \( f_\phi^d \) to produce embeddings \(Z_i^c\) and \(Z_i^d\) from the inputs, and projection heads \( g_{\psi_c}^c \), \( g_{\psi_d}^d \) such that the shared embedding space \( H(Z) \in \mathbb{R}^{D} \) satisfies: (i) a classifier \( h_\omega:Z\to Y \) produces accurate predictions \(\hat{y}_i = h_w(Z)\); (ii) paired embeddings \( \|z_i^c - z_i^d\| \) are optimized for modality invariance; and (iii) the embedding \( z \) is decorrelated from sensitive attribute \( s \) for fairness. The overall architecture is illustrated in Figure~\ref{figArchitecture}.

    \subsection{Overall Architecture}
\begin{figure}[t!]
        	\centering
        	\includegraphics[width=.85\textwidth]{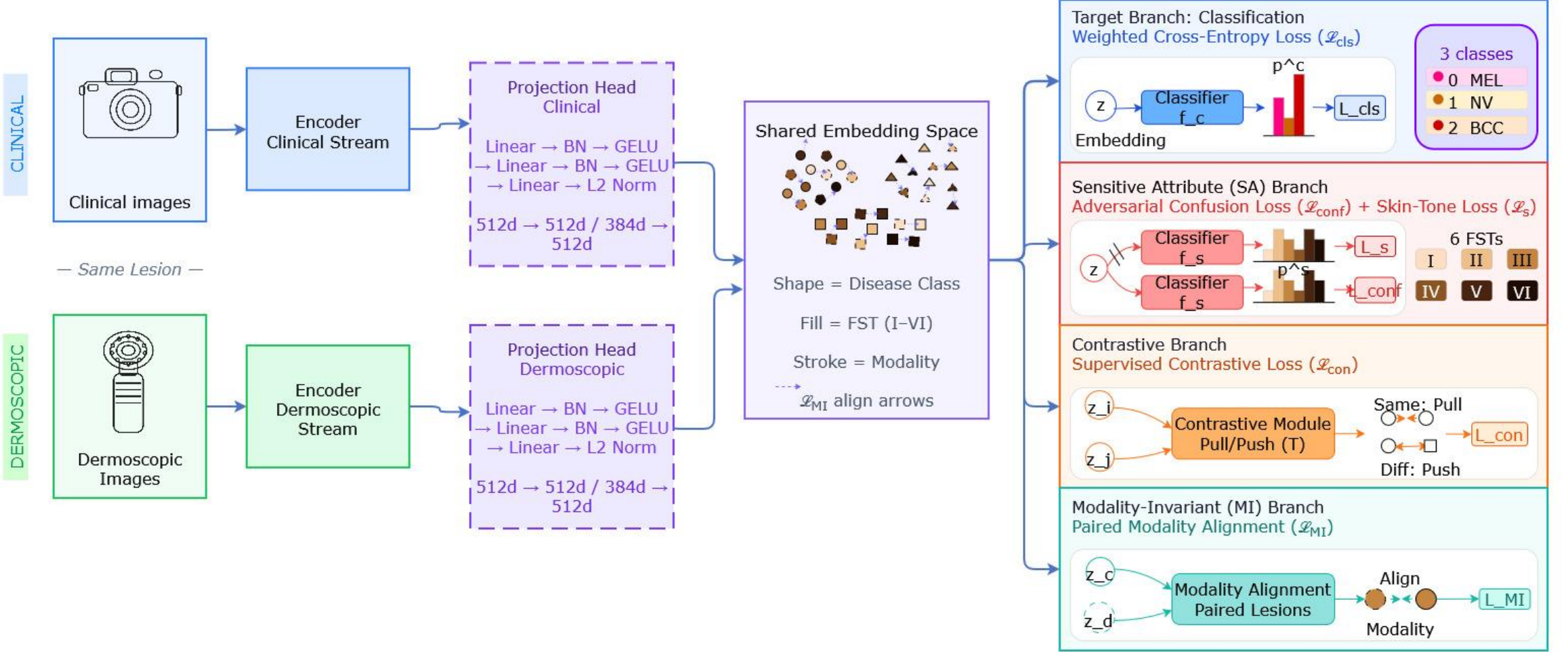}
        	\caption{\textbf{Overview of the proposed architecture}. Clinical and dermoscopic streams are processed by independent encoders and projected into a shared embedding space via modality-specific projection heads. Four training branches jointly optimize classification (\(L_{\text{cls}}\)), skin-tone fairness via adversarial confusion and skin-tone losses (\(L_{\text{conf}}\), \(L_{s}\)), cross-modality contrastive alignment (\(L_{\text{con}}\)), and paired modality invariance (\(L_{\text{MI}}\)).}
        	\label{figArchitecture}
        \end{figure}

        The clinical encoder \( f_\theta^c \) and dermoscopic encoder \( f_\phi^d \) independently process %\(224\times224\) 
        input images (Fig.\,\ref{figArchitecture}). Each is followed by a modality-specific projection head; a simple multi-layer perceptron with batch normalization and GELU activations, ending with L2 normalization, that maps the encoder outputs into a shared \(D\)-dimensional embedding space where the cosine distance between embeddings directly reflects their semantic similarity. These projection heads have independent parameters, producing \( \mathbf{z}_i^c = g_{\psi_c}^c(f_\theta^c(x_i^c)) \) and \( \mathbf{z}_i^d = g_{\psi_d}^d(f_\phi^d(x_i^d)) \in \mathbb{R}^{512} \), meaning the embeddings are converted to a $D$-dimension. The shared embedding space \(Z\) is where all branches contribute to the final classification.
        %\hl{Projection head explained better}

    \subsection{Training Branches and Loss Functions}
        %Four parallel branches structure the embedding space: the target branch, the skin-tone attribute branch, the constrastive and multimodal invariant branches.
        The embedding space is structured by four parallel branches: target classification, skin-tone attribute, contrastive, and modality-invariant.

        \paragraph{\textbf{Target branch (BASE).}} The weighted cross-entropy loss used inverse-frequency class weights  (\(w_{y_i}\propto 1/n_{y_i}\)) to address class imbalance:
        \begin{equation}
            \mathcal{L}_{\text{cls}} = -\frac{1}{N}\sum_{i=1}^{N} w_{y_i} \log p(y_i\mid z_i). 
            \label{eq:cls}
        \end{equation}

        \paragraph{\textbf{Skin-tone Attribute (SA) branch.}} This branch removes skin-tone bias from the learned features. An auxiliary classifier \( f_s\) is attached  to predict skin tone from those features. Training involves two competing losses that pull in opposite directions \cite{duFairDisCoFairerAI2022}. First, the confusion loss \(\mathcal{L}_{\text{conf}}\) updates only the main encoder by removing skin-tone information from the representations. It forces the classifier's predictions to become as uncertain as possible, and is minimized when the classifier outputs equal probability for all six skin-tone classes. Second, the skin loss \(\mathcal{L}_{\text{skin}}\) updates only the classifier. It trains the classifier to predict the true skin-tone label from the current features. This prevents the classifier from collapsing and ensures it continues providing useful feedback. 
        
        Together, these losses create an adversarial game: the encoder tries to hide skin-tone cues, while the classifier tries to detect them. Over time, the encoder learns features that are useful for disease diagnosis but contain no skin-tone information. This adversarial disentanglement mechanism is computed as:
        \begin{equation}
            \mathcal{L}_{\text{conf}} = -\frac{1}{N}\sum_{i=1}^{N}\log(p_s^i), \quad
            \mathcal{L}_{\text{skin}} = -\sum_{k=1}^{6}\log f_s(z)_k.
            \label{eq:sa}
        \end{equation}
        \paragraph{\textbf{Contrastive branch.}} A supervised contrastive loss pulls same-class embeddings together and pushes different classes apart \cite{duFairDisCoFairerAI2022}, computed separately per modality to establish intra-class compactness before cross-modal alignment:
        \begin{equation}
            \mathcal{L}_{\text{contr}} = -\frac{1}{|P_y|}\sum_{p\in P_y} \log\frac{\exp(\Psi(r,p)/\tau)}{\exp(\Psi(r,p)/\tau)+\sum_{n\in N_y}\exp(\Psi(r,n)/\tau)},
            \label{eq:contr}
        \end{equation}
        where \(r=H(Z)\) is the shared embedding space, \(p\) and \(n\) are the distances between an embedding and another embedding in the same class, and in a different class respectively, \( \Psi (.,.) \) is cosine similarity and \( \tau>0 \) a temperature.

        \paragraph{\textbf{Modality-invariant branch.}} 
        This is our core contribution to enforce modality invariance without requiring annotations. The branch employs a VICReg-inspired loss \cite{bardesVICRegVarianceInvarianceCovarianceRegularization2022} with three complementary terms:
        \begin{itemize}
            \item \textbf{\textit{Invariance term} \(\mathcal{L}_{\text{inv}}\)} minimizes the cosine distance between paired clinical and dermoscopic embeddings, forcing the representation to be determined by disease content rather than imaging modality:
            \begin{equation}
                \mathcal{L}_{\text{inv}} = \frac{1}{|P|}\sum_{(i,j)\in P}\left(1 - \frac{z_i^c\cdot z_j^d}{\|z_i^c\|\|z_j^d\|}\right).
                \label{eq:inv}
            \end{equation}
            \item \textbf{\textit{Variance term} \(\mathcal{L}_{\text{var}}\)} prevents dimensional collapse by ensuring that each embedding dimension retains sufficient variance across the batch, applied separately to clinical and dermoscopic streams:
            \begin{equation}
                \mathcal{L}_{\text{var}} = \frac{1}{2d}\sum_{m\in\{c,d\}}\sum_{k=1}^{d}\max\left(0,\gamma - \sqrt{\text{Var}(z_{.k}^m)+\epsilon}\right),
                \label{eq:var}
            \end{equation}
            where \(\gamma=1.0\) is a target variance floor and \(\epsilon\) a small stabilizer.
            \item \textbf{\textit{Covariance term} \(\mathcal{L}_{\text{cov}}\)} decorrelates embedding dimensions to reduce redundancy and increase effective capacity:
            \begin{equation}
                \mathcal{L}_{\text{cov}} = \frac{1}{2d}\sum_{m\in\{c,d\}}\sum_{i\neq j}[C^m]_{ij}^2,
                \label{eq:cov}
            \end{equation}
            where \( C^m = \frac{1}{N-1}(Z^m)^\top Z^m \) is the batch covariance matrix.
        \end{itemize}
        The total MI loss combines these terms:
        \begin{equation}
            \mathcal{L}_{\text{MI}} = \lambda_{\text{inv}}\mathcal{L}_{\text{inv}} + \lambda_{\text{var}}\mathcal{L}_{\text{var}} + \lambda_{\text{cov}}\mathcal{L}_{\text{cov}}.
            \label{eq:mi_total}
        \end{equation}
        The complete objective jointly optimizes all components:
        \begin{equation}
            \mathcal{L}_{\text{Total}} = \mathcal{L}_{\text{cls}} + \lambda_{\text{conf}}\mathcal{L}_{\text{conf}} + \lambda_{\text{skin}}\mathcal{L}_{\text{skin}} + \lambda_{\text{contr}}\mathcal{L}_{\text{contr}} + \lambda_{\text{MI}}\mathcal{L}_{\text{MI}}.
            \label{eq:total}
        \end{equation}
        %At inference, auxiliary branches are discarded; only encoders, projection heads, and the target branch are retained, enabling testing and deployment on either clinical or dermoscopic modalities, or both at a time.
        At inference, the auxiliary branches are discarded, retaining only the encoders, projection heads, and target classification branch. Compared to baselines (FairDisCo and PatchAlign), the resulting model can directly operate on clinical images, dermoscopic images, or both, enabling flexible deployment in resource-constrained settings where only a single imaging modality may be available, thereby eliminating the need for separate specialized models for each modality.

\section{Experiments}
    \subsection{Datasets}
        \begin{table}[t]
        \caption{Dataset statistics after filtering to the target classes (MEL, NV, BCC).}
        \vspace{0.2cm}
        \centering    
        \begin{tabular}{|l|c|c|c|c|c|}
            \hline
            \textbf{Dataset} & \textbf{\# Samples} & \textbf{Retained} & \textbf{Melanoma} & \textbf{Nevus} & \textbf{BCC} \\
            \hline
            HIBA \cite{riccilaraDatasetSkinLesion2023} & 1,616 & 1,195 & 253 & 602 & 340 \\
            Derm7pt \cite{kawahara7PointChecklist2019} & 2,022 & 1,738 & 504 & 1,150 & 84 \\
            PAD-UFES-20 \cite{pachecoPADUFES20SkinLesion2020} & 2,098 & 1,141 & 52 & 244 & 845 \\
            ISIC 2019 \cite{combaliaDermoscopyQualityAssessment2019} & 33,769 & 25,517 & 5,849 & 15,370 & 4,298 \\
            \hline
        \end{tabular}
        \label{tab:datasets}
    \end{table}

    We classify the images into three classes: Melanoma (MEL), Nevus (NV), and Basal Cell Carcinoma (BCC). The training dataset consisted of two paired datasets (Table\,\ref{tab:datasets}). HIBA \cite{riccilaraDatasetSkinLesion2023} has 1,195 images with FST annotations. Its class proportions are 21.2\% MEL, 50.4\% NV, and 28.4\% BCC. Derm7pt \cite{kawahara7PointChecklist2019} has 1,738 images with FST estimated using the Individual Typology Angle, following the implementations of \cite{grohEvaluatingDeepNeural2021}. Its class proportions are 29.0\% MEL, 66.2\% NV, and 4.8\% BCC. For both datasets, only Melanoma, Nevus, and Basal Cell Carcinoma classes were retained, focusing on the most clinically relevant and well-represented classes. Data were stratified by disease class and FST. We split it into 70\% training, 10\% validation, and 20\% test. All images were resized to $224 \times 224$ pixels and online augmentations were applied, including random resized crop, color jitter, and horizontal flip with a given probability ($p=0.85$). For cross-dataset evaluation, we used two external benchmarks. PAD-UFES-20 \cite{pachecoPADUFES20SkinLesion2020} contains 1,141 clinical images with FST labels. Its class proportions are 4.6\% MEL, 21.4\% NV, and 74.0\% BCC. ISIC 2019 \cite{combaliaDermoscopyQualityAssessment2019} has 25,517 dermoscopic images. Its class proportions are 22.9\% MEL, 60.2\% NV, and 16.8\% BCC.

    \subsection{Experimental Setup}
        \paragraph{\textbf{Model training and hyperparameters.}} Based on preliminary experiments, the ViT-Small (patch16-224) model pre-trained on ImageNet was adopted as backbone encoder. Key hyperparameters were selected on the validation split, with the final configuration balancing %%%\hl{the task-specific objectives}. 
        %Critically, \hl{two loss weights were fixed} to $1.0$ and the other weights were \hl{adjusted and vice versa}. 
        The loss weights were set to $\lambda_{\text{conf}}=0.3$, $\lambda_{\text{skin}}=0.2$, and $\lambda_{\text{contr}}=0.5$, prioritizing the contrastive alignment over skin classification. 
        The modality-invariant term was assigned a smaller weight $\lambda_{\text{MI}}=0.15$ after adjusting, since a larger value would have overly suppressed task-relevant variations. For the sub-weights of the variance-invariance-covariance regularizer, following \cite{bardesVICRegVarianceInvarianceCovarianceRegularization2022}, we kept $\lambda_{\text{inv}}=1.0$ and $\lambda_{\text{var}}=1.0$ to enforce strict invariance and variance preservation, while setting $\lambda_{\text{cov}}=0.04$ to apply only mild decorrelation pressure. The temperature $\tau=0.1$ follows standard contrastive practice, and the variance floor $\gamma=1.0$ ensures numerical stability. A batch size of $32$ was used due to memory constraints, and AdamW optimizer \cite{kingmaAdamMethodStochastic2017} with separate learning rates of $8.5\times10^{-5}$ for the encoders and $1\times10^{-4}$ for the classifiers and projection heads. Weight decay of $1\times10^{-4}$ and $50$ warmup epochs were used to stabilize early training, while label smoothing of $0.01$ mitigates overfitting to noisy labels. All models were trained for $500$ epochs on the MatriCS platform using one Nvidia GPU, 16 CPU cores, and 128GB of host memory.

        \paragraph{\textbf{Model Evaluation.}} Our model is compared with two main baseline: FairDisCo \cite{duFairDisCoFairerAI2022} and PatchAlign \cite{aayushmanFairAccurateSkin2024}. The predictive performance is assessed via standard classification metrics: Accuracy (Acc), F1-score, and Area Under the ROC Curve (AUC). Fairness is evaluated following \cite{duFairDisCoFairerAI2022} and \cite{kongAchievingFairnessChannel2024} by computing: Predictive Quality Disparity (PQD) as the ratio of minimum to maximum class-wise accuracy across skin tones; and Accuracy Gap (\(\Delta\)Acc) as the difference between highest and lowest group accuracies.

\section{Results}
    \subsection{Ablation Study}
        We first perform ablation experiments on the internal test sets to isolate the contribution of each loss component (Tab.~\ref{tab:combined_all}). Starting from a dual-encoder baseline with classification loss only (BASE), we progressively added the skin-tone attribute branch (SA), contrastive branch (Contr), and Modality-Invariant branch (MI). On the internal test set, SA alone improves accuracy from 93.13\% at the BASE, to 93.52\%, and reduces the Acc gap from 0.0691 to 0.0562. MI alone improves the BASE PQD from 66.67\% to 87.23\%. The combination BASE+Contr+MI yields the lowest Acc gap (0.0265), suggesting contrastive learning with modality invariance implicitly promotes fairness even without explicit SA constraints. Removing MI from the full model increases Acc gap from $0.054$ to $0.0747$, confirming that modality invariance provides complementary fairness regularization.

    \subsection{Comparison with State-of-the-Art}
        \paragraph{\textbf{HIBA+Derm7pt internal test set.}}Table~\ref{tab:combined_all} compares the full model against FairDisCo and PatchAlign. Our model achieves 93.33\% accuracy and 92.17\% F1-score, which does not surpass but remains highly competitive with FairDisCo (93.81\%) and PatchAlign (93.81\%). Crucially, our model reaches a PQD of 87.23\%, substantially surpassing FairDisCo (66.67\%) and PatchAlign (82.98\%).

        \paragraph{\textbf{PAD-UFES-20 external benchmark.}} The full model consistently outperforms state-of-the-art methods in Acc (85.10\%), AUC (91.89\%), and F1-score (66.25\%). This clearly surpasses FairDisCo (78.35\% Acc, 83.44\% AUC) and PatchAlign (75.02\% Acc, 88.66\% AUC). Our model prioritizes overall diagnostic accuracy and robust cross-dataset generalization, validating its real-world applicability although PatchAlign exhibits a higher PQD (0.8878) than ours (0.54).
        This fairness gap likely reflects distributional shift: PAD-UFES-20 is heavily skewed toward BCC (74\% vs. 28.4\% and 4.8\% in HIBA and Derm7pt), which may interact unevenly with underrepresented skin-tone subgroups.

        \paragraph{\textbf{ISIC 2019 external benchmark.}} The full model achieves 76.64\% Acc and 70.62\% F1-score (matching FairDisCo and surpassing PatchAlign), while its AUC (80.50\%) is significantly lower than BASE (86.62\%). This is a predictable trade-off for modal invariance: the dermoscopic encoder regularized to align with clinical embeddings sacrifices some modality-specific discrimination for cross-modality generalizability, a modest cost given the substantial gain achieved on PAD-UFES-20. Fairness metrics were not computed on ISIC 2019 because, as a purely dermoscopic dataset, skin surface is not visible in the images, making skin tone difficult to estimate, therefore impacting fairness evaluation.

\begin{table}[t!]
    \centering
    \caption{Results on internal and external datasets: ablations and state-of-the-art comparison. The best values are in \textbf{bold}, and the second are \underline{underlined}}\label{tab:combined_all}
    \vspace{0.2cm}
    \scriptsize
    \begin{tabular}{|l|l|l|l|l|l|l|}
        \hline
        \multirow{2}{*}{\textbf{Dataset}} & \multirow{2}{*}{\textbf{Model}} & \multicolumn{3}{l|}{\textbf{Performance}} & \multicolumn{2}{l|}{\textbf{Fairness}} \\
        \cline{3-7}
         & & \textbf{Acc $\uparrow$} & \textbf{AUC $\uparrow$} & \textbf{F1-Score $\uparrow$} & $\Delta$\textbf{Acc $\downarrow$} & \textbf{PQD $\uparrow$} \\
        \hline
        \multirow{10}{*}{HIBA+Derm7pt} & BASE & .9313 & .9738 & .9165 & .0691 & .6667 \\
        & BASE+SA & .9352 & \textbf{.9813} & .9226 & .0562 & .6667 \\
        & BASE+Contr & .9284 & .9683 & .9142 & .0887 & .8298 \\
        & BASE+MI & .9188 & .9619 & .9001 & \underline{.0394} & \underline{.8723} \\
        & BASE+Contr+MI & .9043 & .9664 & .8883 & \textbf{.0265} & .8511 \\
        & BASE+SA+MI & .9236 & .9560 & .9053 & .0546 & \textbf{.8871} \\
        & BASE+SA+Contr & .9333 & .9748 & .9186 & .0747 & .8511 \\
        & FairDisCo & \textbf{.9381} & .9790 & \textbf{.9268} & .0769 & .6667 \\
        & PatchAlign & \textbf{.9381} & \underline{.9792} & \underline{.9244} & .0988 & .8298 \\
        & \textbf{All} & .9333 & .9678 & .9217 & .0540 & \underline{.8723} \\
        \hline
        \multirow{4}{*}{PAD-UFES-20} & BASE & .7528 & .8404 & .5040 & .2066 & .6053 \\
        & FairDisCo & .7835 & .8344 & .5421 & .1479 & .5631 \\
        & PatchAlign & .7502 & .8866 & .5935 & \underline{.0760} & \textbf{.8878} \\
        & \textbf{All} & \textbf{.8510} & \textbf{.9189} & \textbf{.6625} & .1413 & .5400 \\
        \hline
        \multirow{4}{*}{ISIC 2019} & BASE & \underline{.7586} & \textbf{.8662} & .6998 & / & / \\
        & FairDisCo & .7560 & .8335 & \underline{.7050} & / & / \\
        & PatchAlign & .7337 & \underline{.8626} & .6650 & / & / \\
        & \textbf{All} & \textbf{.7664} & .8050 & \textbf{.7062} & / & / \\
            \hline
        \end{tabular}
        % \vspace{3pt}
    \end{table}

    \subsection{Embedding Space Visualizations}
\begin{figure}[t!]
    	\centering
    	\includegraphics[width=.74\textwidth]{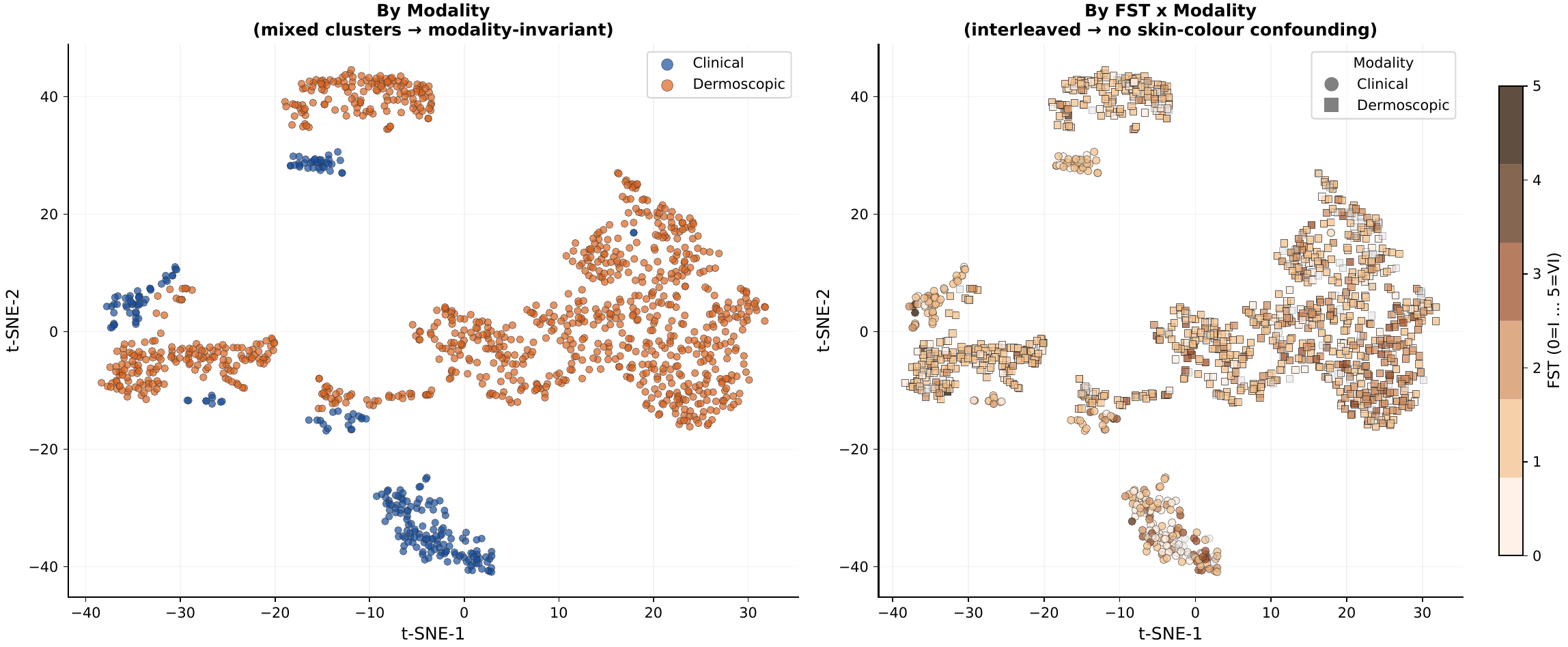}
    	\caption{Left: t-SNE of the full model colored by imaging modality, showing alignment of clinical and dermoscopic embeddings. Right: t-SNE colored by Fitzpatrick skin type, showing near-perfect skin-tone mixing within disease clusters.}
    	\label{figTsneClassVit}
        \vspace{-0.3cm}
    \end{figure}

    We produced t-SNE plots of the full model on the internal test set, making it easier to understand how the shared embedding space is organized in 2D scatter plots and how alignment takes place. With these considerations in mind, the resulting visualizations (Fig.~\ref{figTsneClassVit}) reveal interleaved clinical-dermoscopic embeddings (confirming modality invariance), and near-perfect mixing of skin tones within each cluster (confirming fairness).

\section{Discussion}
    The dual ViT-Small design has about 45M parameters, which limited the batch size to $32$ and needed separate learning rates for the encoders and the classifiers. Among the five loss weights, the three MI sub-weights ($\lambda_{\text{inv}}$, $\lambda_{\text{var}}$, $\lambda_{\text{cov}}$) were carefully tuned: a small covariance penalty %($\lambda_{\text{cov}}=0.04$)
    reduced redundancy without hurting quality, while high invariance and variance %$\lambda_{\text{inv}}=\lambda_{\text{var}}=1.0$ 
    enforced strong invariance and prevented collapse. 
    Because the five weights interact, they were tuned one at a time on the validation split rather than searching all combinations, which is practical but does not guarantee optimal performance. A more principled way for hyperparameter selection could lead to better performance while reducing complexity.
    Fairness supervision also relies on estimated Fitzpatrick skin types for Derm7pt~\cite{grohEvaluatingDeepNeural2021}, adding some label noise, and the AUC drop in ISIC 2019 shows that modality invariance costs some modality-specific detail, which was a trade-off given the fairness and generalization gains. 
    Finally, all datasets come from International repositories, with no African skin lesion data, reflecting the broader scarcity of paired, skin-tone-labeled dermatology data from African populations. This gap is particularly relevant given that public dermatology datasets remain skewed toward lighter Fitzpatrick types (I–III)~\cite{daneshjouDisparitiesDermatologyAI2022,grohEvaluatingDeepNeural2021}, while African populations are concentrated in FST IV–VI, the range where prior fairness methods have historically underperformed. Adapting MIFR to African clinical datasets, such as eSkinHealth \cite{wang2025eskinhealth}, represents an important direction for future work.

    \section{Conclusion}
    This work proposed a Modality-Invariant and Fair Representation Learning (MIFR) framework for equitable skin disease classification across clinical and dermoscopic imaging. The model demonstrates superior cross-dataset generalization, achieving an AUC improvement of over 3.2\% on PAD-UFES-20 compared with the best-performing baseline (PatchAlign) and over 7.8\% compared with the base model (BASE), while achieving substantial fairness gains on the internal dataset and maintaining competitive predictive performance across datasets. t-SNE visualizations confirm modality-aligned embeddings, %with suppressed skin-tone information, 
    validating that joint optimization of invariance and fairness yields more robust and equitable models. However, MIFR has two key limitations. First, it relies on paired training data, which are often scarce. Second, its dual-encoder architecture increases computational cost. Future work should investigate unpaired approaches to modality invariance representation learning and develop more efficient architectures to facilitate deployment in resource-constrained settings.

\begin{credits}
    \subsubsection{\ackname}
        This work was granted access to HPC resources of “Plateforme MatriCS” within the University of Picardie Jules Verne. “Plateforme MatriCS” is co-financed by the European Union with the European Regional Development Fund (FEDER) and the Hauts-De-France Regional Council among others. We thank the Hertie Foundation and the Excellence Cluster 2064 “Machine Learning – New Perspectives for Science” for supporting this work.

    \subsubsection{\discintname}
        The authors declare no competing interests.
\end{credits}

\bibliographystyle{splncs04}
\bibliography{bibliography}

\end{document}